# Image-Based Fire Detection in Industrial Environments with YOLOv4

Otto Zell[1], Joel Pålsson[1], Kevin Hernandez-Diaz[1][a], Fernando Alonso-Fernandez[1][b], Felix Nilsson[2]
[1]*School of Information Technology (ITE), Halmstad University, Sweden*
[2] *HMS Industrial Networks AB, Halmstad, Sweden*
ottozell0023@gmail.com, joelpalsson1998@hotmail.com{kevher, feralo}@hh.se, fenil@hms.se

Keywords: Fire detection, Smoke Detection, Machine learning, Computer Vision, YOLOv4

Abstract: Fires have destructive power when they break out and affect their surroundings on a devastatingly large scale. The best way to minimize their damage is to detect the fire as quickly as possible before it has a chance to grow. Accordingly, this work looks into the potential of AI to detect and recognize fires and reduce detection time using object detection on an image stream. Object detection has made giant leaps in speed and accuracy over the last six years, making real-time detection feasible. To our end, we collected and labeled appropriate data from several public sources, which have been used to train and evaluate several models based on the popular YOLOv4 object detector. Our focus, driven by a collaborating industrial partner, is to implement our system in an industrial warehouse setting, which is characterized by high ceilings. A drawback of traditional smoke detectors in this setup is that the smoke has to rise to a sufficient height. The AI models brought forward in this research managed to outperform these detectors by a significant amount of time, providing precious anticipation that could help to minimize the effects of fires further.

## 1 INTRODUCTION

The use of smoke detectors have become standard on fire detection (Ahrens, 2021). They are excellent at quickly detecting fires in regular households. There are, however, still scenarios where a smoke detector could be improved upon. Buildings with large volumes of non-fire smoke could cause false alarms, causing the fire brigade to waste resources and possible penalties for the user. Likewise, warehouses with ceilings high above the floor struggle with smoke not reaching the detectors fast enough.

There are several types of detectors, each one with different benefits and drawbacks (Accosta and Martin, 2017). Point smoke detectors, carbon monoxide detector,s and aspirating detection systems all have a general applicable maximum ceiling height of 10.5 meters. Optical beam smoke detectors have a maximum ceiling height of 25 meters, and 40 if configured in extra sensitive mode. Optical beam smoke detectors can detect fire in a very high ceiling but require the smoke to rise to the height at which the detectors are installed. This type of alarm also favors medium-growth fires compared to fast-growth fires.

This research explores using neural networks to detect fire and smoke in real time using images from a camera stream. We do not aim at replacing smoke detectors, but to explore the potential of Artificial Intelligence (AI) for smoke and fire detection. This AI would potentially work as an extra layer of protection, integrated with a smoke detector for added security in environments where the smoke detector might be less efficient (Bu and Gharajeh, 2019).

This project is a collaboration of Halmstad University with HMS Networks AB in Halmstad. HMS offers a portfolio of products for remote control of field equipment and premises (HMS, 2022). They explore emerging technologies for industry, and one crucial technology is AI, where they want to examine and showcase different possibilities and applications of AI and vision technologies, e.g. (Nilsson et al., 2020), which may be part of future products. The datasets and models trained during this research will be made available in our repository[1].

## 2 RELATED WORKS

There are several works related to fire detection using chemical (Fonollosa et al., 2018) and camera-based vision systems (Bu and Gharajeh, 2019). The latter covers different intelligent techniques (like CNNs, color information, etc.) to detect fires in forest fires and general indoor/outdoor environments.

---
[a] https://orcid.org/0000-0002-9696-7843
[b] https://orcid.org/0000-0002-1400-346X

[1]https://github.com/HalmstadUniversityBiometrics/Fire-detection-in-industrial-environments-with-Yolov4

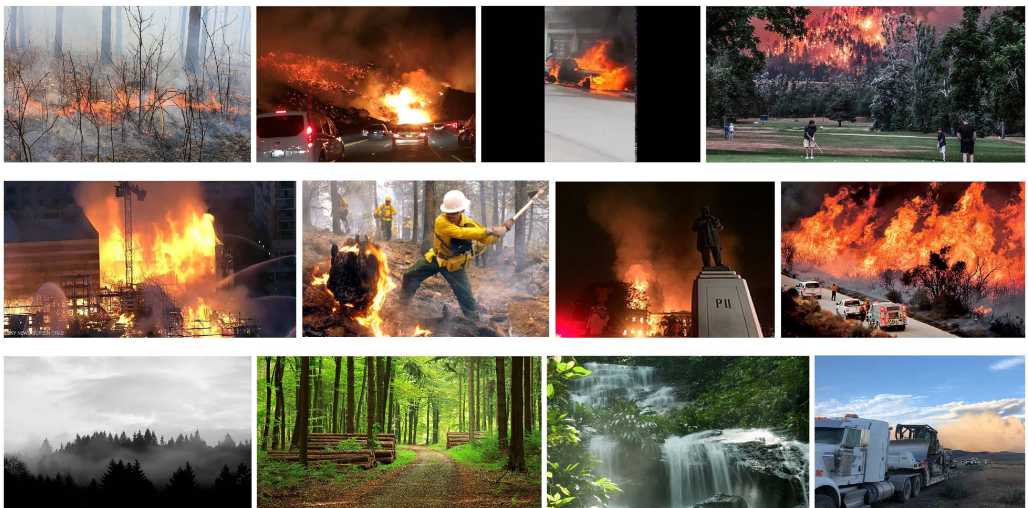

Figure 1: Examples of images from Dataset 2 with positive examples (rows 1,2) and negative examples (row 3).

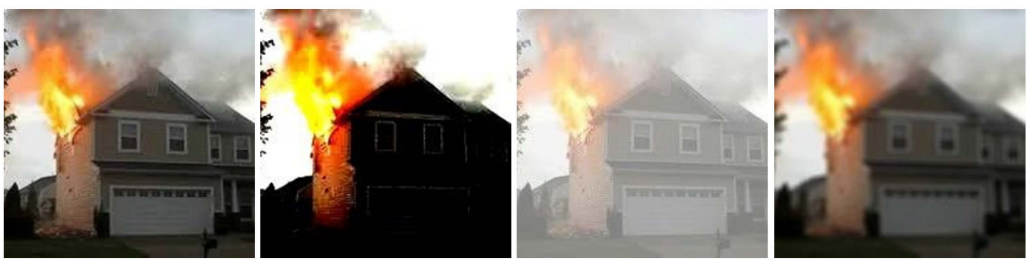

Figure 2: Examples of data augmentation (first image: original, second: brightness, third: contrast, fourth: blur).

(Gonzalez et al., 2017) demonstrated that fires could accurately and quickly be identified using CNNs, allowing real-time operations. It was part of an Unmanned Autonomous Vehicle (UAV) system to identify wildfires from above using drones, and calculate location and distance. The CNN, based on AlexNet, had an accuracy of 94.76%. The model was trained on 450 positive and 50 negative images and tested on 50 positive and 10 negative images representing aerial views of elements that could be misclassified as fire.

(Muhammad et al., 2019) proposed an architecture based on SqueezeNet. The fully-connected layers are removed, and instead, binarization is applied to feature maps of several layers, found to be sensitive to fire regions. This provides the segmented fire, allowing to calculate its size. Passing the image again through the original SqueezeNet provides one of the 1000 classes of ImageNet, which can be used to determine the situation in the scene, such as a fire in a house, a forest, or a vehicle. A total of 68,457 images from public sources were used, with 5258 fire images and 5061 non-fire images used for training.

(Zhang et al., 2021) proposed a new architecture for semantic segmentation where each individual pixel is labeled, an alternative to the classical bounding box-based approach. Semantic segmentation holds much potential as it eliminates the extra noise when drawing square boxes around fires. The model is based on a U-Net segmentation network that uses a modified SqueezeNet as encoder-encoder, and an Attention Gate unit in the skip connections. Experiments are targeted toward forest fires. Data is gathered from public sets, consisting of 1135 fire/500 non-fire images for segmentation experiments, and 565 fire/1990 non-fire for classification experiments. The obtained accuracy is 93%, with an average prediction time of 0.89 seconds per image.

(Abdusalomov et al., 2021) modified YOLOv3 to detect fires in a variety of urban environments. They also applied a variety of data augmentation techniques (rotation, contrast, brightness) to 9,200 nighttime and daytime flame pictures gathered from public sources, obtaining 624,900 images. They also added 20,000 fire-like images to counteract false positives. An interesting observation is that color-based augmentation actually decreased training accuracy, very likely because fire color is an important attribute for its detection. They trained the model for 50,000 iterations, finding that the test accuracy of YOLOv3 (97.8%) is higher than other models based on tiny models (YOLOvX-tiny) or other versions (YOLOv4).

Table 1: Datasets employed to create each model, the number and predominant type of fires on each.

|        | D1 | D2 | D3 | D4 | D5 | #positive | #negative | #total | type                        |
|--------|----|----|----|----|----|-----------|-----------|--------|-----------------------------|
| Model1 | X  |    |    |    |    | 361       | 51        | 412    | candles, houses             |
| Model2 |    |    | X  |    |    | 1444      | 204       | 1648   | candles, houses             |
| Model3 | X  |    | X  |    |    | 361       | 100       | 461    | candles, houses, warehouses |
| Model4 | X  | X  |    |    |    | 1109      | 295       | 1404   | candles, houses, forest     |
| Model5 |    |    |    | X  | X  | 1444      | 400       | 1844   | candles, houses, warehouses |

(Avazov et al., 2022) improved YOLOv4 via data augmentation and modifying the network as well. Based on the wording and paper structure, it seems that the data employed is the same as (Abdusalomov et al., 2021), although the authors are different. The 9,200 source pictures are augmented to 80,400 via rotation, contrast, and brightness modifications. 10,000 fire-like images are also added as negatives. On this occasion, the test accuracy of YOLOv4 (96.3%) is higher than other models based on tiny models (YOLOvX-tiny) or previous versions (YOLOv3).

# 3 METHODOLOGY

## 3.1 Data Acquisition

Data had to be sourced and labeled manually since no pre-existing labeled data could be found. Five data sets were sourced and labeled, which will be referred to as Datasets 1, 2, 3, 4, and 5. Dataset 1 was sourced from (Avazov et al., 2022), with 412 images (51 negative), mostly of candles and houses burning. The mentioned paper indicates 9200 images in total, but in the authors' repository, we only found 412 available images. Dataset 2 was sourced from kaggle.com[2], with 992 images (244 negative), mostly of outdoor fires in forests, and a few houses and cars burning. Negative images are from nature, e.g.: forest, tree, grass, river, people, foggy forest, lake, animal, road, and waterfall. Dataset 3 was sourced from pictures of warehouses from istockphoto.com, containing only 49 negative images. All sets were labeled manually. Data set 4 and 5 are augmented versions of Dataset 1 and 3, respectively, quadrupling the size by changing brightness, contrast, and blur. Brightness was increased by 50%, contrast by 50%, and the picture blurred by 5 with OpenCV. Figure 1 shows some examples from Dataset 2, whereas an example of data augmentation can be seen in Figure 2.

## 3.2 System Overview

This research has developed an AI to detect fires and smoke with a camera. Object detection is a common classification problem that attempts to find and classify specific objects in a frame. Several techniques have been developed in this branch of machine learning. When choosing a model for this research, a couple of requirements had to be met. The most critical was the real-time requirement and the fact that we plan to deploy it on an NVIDIA Jetson nano. This narrows down the choice to one-stage object detectors. Picking a one-stage over a two-stage detector means sacrificing mAP (Mean Average Precision) for speed (Li et al., 2017).

Two models match our requirements: YOLOv4 (Bochkovskiy et al., 2020) and EfficientDet (Tan et al., 2020). EfficientDet falls behind YOLOv4 in speed and mAP (Wu et al., 2020), so it will be our choice for this work. YOLO detectors also have features that other models lack, such as contextual awareness, seeing the entire image during training to obtain a 'bigger picture' and detect objects more accurately because of context. This is advantageous for our purpose (Bochkovskiy et al., 2020), since context matters when it comes to fires. Previous work also suggests that YOLOv4 performs well for real-time fire detection (Avazov et al., 2022).

# 4 EXPERIMENTS AND RESULTS

Five YOLO models have been trained for this paper to detect two classes, fire and smoke. Each model employs a certain combination of the available datasets, as described in Table 1. Data division for training/testing is 80/20%. The training parameters are batch size of 64, resolution=576×576 (RGB images), momentum=0.949, decay=0.0005, learning rate=0.001, burn-in=1000. The models are trained on a machine with an Nvidia RTX 3070 GPU with 8Gb RAM and an AMD Ryzen 7 5700x CPU with 32GB RAM during 3000 iterations. The models are evaluated from an object detection and recognition/classification perspective.

---
[2]https://www.kaggle.com/datasets/phylake1337/fire-dataset

Table 2: Comparison of models from an *object detection* perspective. Bold numbers indicate the best result of each column, whereas underlined numbers indicate the runners-up.

|  | Fire | | | Smoke | | | Fire + Smoke | | | | | |
| --- | --- | --- | --- | --- | --- | --- | --- | --- | --- | --- | --- | --- |
|  | TP | FP | AP | TP | FP | AP | TP | FP | FN | Precision | Recall | F1 |
| Model1 | 236 | 79 | 75.2% | 61 | 28 | 53.2% | 297 | 107 | 117 | <u>74%</u> | <u>72%</u> | 73% |
| Model1A | 205 | 231 | 47.5% | 60 | 27 | 53.2% | 265 | 258 | 142 | 51% | 65% | 57% |
| Model2 | 240 | 63 | 81.2% | 62 | 35 | 58.2% | 302 | 98 | 105 | <u>75%</u> | <u>74%</u> | <u>75%</u> |
| Model2A | 203 | 217 | 57.0% | 62 | 37 | 56.9% | 265 | 254 | 142 | 51% | 65% | 57% |
| Model3 | 245 | 74 | <u>81.7%</u> | 63 | 80 | 52.0% | 308 | 154 | 99 | 67% | **76%** | 71% |
| Model4 | 200 | 141 | 64.8% | 49 | 35 | 42.2% | 249 | 176 | 158 | 59% | 61% | 60% |
| Model5 | 242 | 55 | **82.3%** | 55 | 19 | **59.5%** | 297 | 74 | 110 | **80%** | <u>73%</u> | **76%** |

Table 3: Comparison of models from an *object recognition* perspective. Bold numbers indicate the best result of each column, whereas underlined numbers indicate the runners-up.

|  | Fire | | | | | | | Smoke | | | | | |
| --- | --- | --- | --- | --- | --- | --- | --- | --- | --- | --- | --- | --- | --- |
|  | TP | TN | FP | FN | Precision | Recall | F1 | TP | TN | FP | FN | Precision | Recall | F1 |
| Model 1 | 145 | 28 | 3 | 2 | 97.9 | **98.6** | <u>98.2</u> | 66 | 94 | 3 | 15 | 81.5 | **97.1** | 88.6 |
| Model 2 | 144 | 30 | 1 | 3 | **99.3** | <u>98</u> | **98.6** | 71 | 93 | 4 | 10 | <u>94.7</u> | 87.6 | **91.0** |
| Model 3 | 141 | 28 | 3 | 6 | 97.9 | 95.9 | 96.9 | 76 | 87 | 10 | 5 | 88.4 | <u>93.8</u> | **91.0** |
| Model 4 | 136 | 25 | 6 | 11 | 95.8 | 92.5 | 94.1 | 68 | 82 | 15 | 13 | 81.9 | 84 | 82.9 |
| Model 5 | 141 | 29 | 2 | 6 | <u>98.6</u> | 95.9 | 97.2 | 64 | 96 | 1 | 17 | **98.4** | 79 | 87.6 |

*Object detection* is defined as the model detecting all individual objects in the picture, e.g. if there are 5 fires, they are all detected. A True Positive (TP) is when the Intersection Over Union (IoU) of the predicted and ground-truth bounding boxes is greater than a pre-defined threshold (in this paper, we use 0.5) and the class is correctly attributed. A False Positive (FP) is when the IoU is below the threshold. A False Negative (FN) is when there is no predicted bounding box over a ground-truth box, or the predicted bounding box is attributed to another class. Table 2 gives the results of our experiments. Precision measures the proportion of true positives of a class that are correct, computed as P=TP/(TP+FP). Recall measures the proportion of actual positives of a class that were predicted correctly, computed as R=TP/(TP+FN). High precision but low recall means that the majority of positive classifications are positives, but the model detects only some of the positive samples (in our case, it would mean missing many fires or smokes). Low precision but high recall means that the majority of positive samples are classified correctly (i.e. few fires or smokes are missed), but there are also too many false positives (i.e. fire or smoke detections that are not true). A single metric that summarizes both parameters is the F1 score, computed as F1=2×(P×R)/(P+R). A high F1 score means that both P and R are high, and vice-versa. Varying the IoU threshold, a precision-recall curve can be obtained, showing the trade-off between the two metrics for different thresholds. The Average Precision (AP) is the area under the precision-recall curve, providing a single metric that does not consider the selection of a particular threshold.

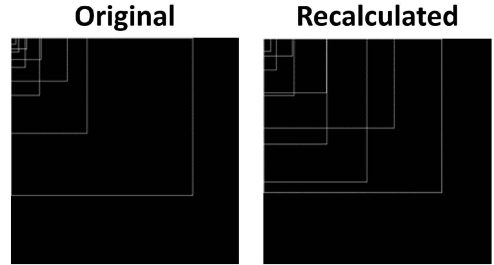

Figure 3: Original (left) and recalculated (right) anchors boxes shown on a 576×576 canvas.

From Table 2, it can be seen that Model5 is generally the best, followed by Model2. These are the models whose databases contain the highest amount of images (Table 1), so this factor could be playing a role. However, Model4, which uses a similar number of images, usually shows the worst performance, the only difference being that it contains images of outdoor fires in the forest, in addition to candles and burning houses. The A after Model1 and Model2 refer to recalculated anchor boxes being used. Recalculations were done with the inbuilt anchor recalculation algorithm provided in the darknet repository. The algorithm uses $k$-means clustering to estimate the ideal anchor boxes. In Figure 3, we show the original and recalculated anchor boxes, with the recalculated being generally larger. We use the default number of 16 boxes. Increasing the number of boxes could potentially increase accuracy but at the cost of execution time. From the results, it can be seen that the models with recalculated anchors generally perform worse (in fire or fire+smoke numbers). In smoke detection, they do not show a significant difference. Our interpretation is that the recalculated anchor boxes are larger due to the smoke generally taking up much of the picture's area compared to fires. As a result, smoke detection is not affected by the new anchor boxes, while fires are strongly affected since they do not necessarily take a portion of the image as large as the smoke. An alternative would be to recalculate the anchors separately for each class. From Table 2, we also observe that fire detection is more accurate than smoke detection in our experiments. This could pose a challenge e.g. if smoke is visible earlier than the flames.

*Object recognition* or classification is defined as the model correctly labeling a picture with fire if it really contains fire (one or several), e.g. on an image with 5 fires, if at least one is detected, it is counted as a TP. A FP is when a box is placed over no fire. A FN is when there is a fire, but no fire is detected. A TN is when an image without fire or smoke is labeled as such. Table 3 gives the results from an *object recognition* perspective. Here, Model2 and Model1 come out on top (in terms of F1 score), although Model5 comes next at a short distance. Thus, Model2 works well for both the object detection and recognition perspectives. Again, fire detection appears as more accurate than smoke detection. If we concentrate on another measures, Model1 shows the best recall, although its precision is not among the best. A high recall means that this model would miss fewer fires, which is more critical, at the cost of more false positives. Another commonality is that Model4 is the worst model in object recognition as well, despite using a similar amount of training and test data than Model2. Model4 is built from Datasets 1 and 2, whereas Model2 is built from Dataset 4, which is an augmented version of Dataset 1 (Table 1). Data augmentation, as carried out in Dataset 4, seems to have a more positive effect than combining different datasets.

When comparing the object detection and object recognition perspectives (Table 2 vs Table 3), it can be observed that object detection (finding individual fires or smoke in a picture) is much more difficult than object recognition (marking if an image has fire or smoke). The latter may be sufficient to just raise an alarm as soon as some fire or smoke has been detected. In Table 4, we show the results of this paper together with other approaches as reported in the related literature. We report our Model2 (object recognition perspective). Although the different methods are not fully comparable due to the use of different databases or protocols, our approach shows comparatively a very competitive performance, with an F1 score of 98.6%. A difference that must be mention though is that all compared approaches have the goal of forest fire prevention, with data coming from satellite pictures. In addition, the datasets of some of those papers are not public (Avazov et al., 2022; Abdusalomov et al., 2021), which means that replication is difficult.

To conclude, we carried out a visual evaluation of our model applied to a video showing fires purposely started inside a warehouse. The frames of such video are not included in the datasets gathered for this paper. A few snapshots are shown in Figure 4. The first part of the video shows a concealed fire behind wooden boxes. Our system detected the first indications of fire at 4:23, whereas the fire alarm at the ceiling activated at 4:35, meaning a 12 seconds gain. In the second part of the video, the fire is just concealed behind some boxes, and our method starts to pick it up very quickly when the flames have not yet reached too much height (at 9 seconds). The ceiling alarm on the left does not catch it up until the flames reach a sufficient height, which happens after 32 seconds, meaning a 23 seconds gain with the camera-based system. The ceiling alarm on the right part takes even more, catching the fire after 43 seconds.

## 5 CONCLUSIONS

This work studies the use of camera-based algorithms for fire and smoke detection in industrial environments like warehouses. For such purpose, we have gathered and labeled appropriate data, and evaluated several models based on the popular YOLOv4 architecture (Bochkovskiy et al., 2020). Our method achieves a competitive accuracy compared to other solutions in the literature, although they are mostly focused on detecting forest fires. The amount and relevance of the training data could be improved, since most of it consists of candles, houses and forest fires. But even with this consideration, the trained YOLO models grasped the concept of fire and generalized well over unseen data, as observed in Figure 4. Our approach performs well in high ceiling environments, showing a substantial decrease in detection time. The improvement in detection time over regular smoke alarms is expected to scale with the ceiling height, since fire detectors must wait for the smoke to rise (Accosta and Martin, 2017). Sufficient coverage with several cameras would also reduce the likelihood of concealed fires that remain undetected, such as the one in the top part of Figure 4.

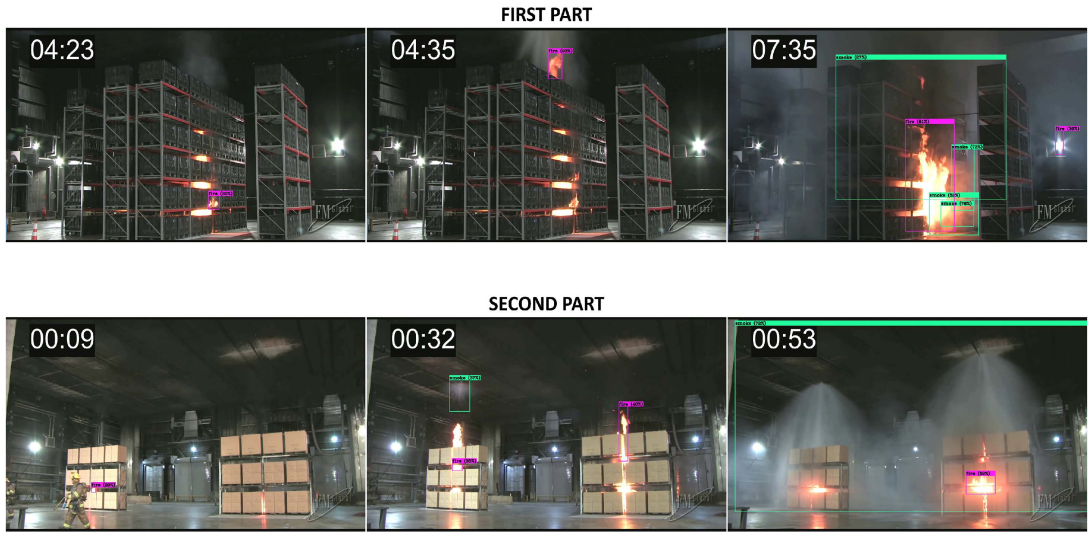

Figure 4: Snapshots of our method applied to a video with fire and smoke. The full video is available at the url: https://github.com/HalmstadUniversityBiometrics/Fire-detection-in-industrial-environments-with-Yolov4.

Table 4: Comparison to other works in the literature.

| Source | Algorithm | Precision | Recall | F1 |
|---|---|---|---|---|
| (Muhammad et al., 2019) | SqueezeNet | 86.0 | 97.0 | 91.0 |
| (Abdusalomov et al., 2021) | AlexNet | 82.0 | 98.0 | 75.1 |
| (Abdusalomov et al., 2021) | Elastic-YOLOv3 | 98.5 | 96.9 | 97.7 |
| (Abdusalomov et al., 2021) | YOLOv3-incremental | 97.9 | 91.2 | 94.3 |
| (Abdusalomov et al., 2021) | Faster R-CNN | 81.7 | 94.5 | 87.2 |
| (Abdusalomov et al., 2021) | Dilated CNNs | 98.9 | 97.4 | 98.2 |
| (Abdusalomov et al., 2021) | ResNet | 94.8 | 93.6 | 94.2 |
| (Abdusalomov et al., 2021) | VGG16 | 97.5 | 87.9 | 98.0 |
| (Abdusalomov et al., 2021) | YOLOv5 | 98.5 | 96.7 | 98 |
| (Abdusalomov et al., 2021) | YOLOv3+OHEM | 86.6 | 77.8 | 89.2 |
| (Abdusalomov et al., 2021) | YOLOv4 | 95.9 | 96.7 | 88.3 |
| (Abdusalomov et al., 2021) | YOLOv3 improved | 98.1 | 99.2 | 99.5 |
| (Avazov et al., 2022) | YOLOv4 | 98.2 | 99.7 | 99.7 |
| ours | YOLOv4 | 99.3 | 98.0 | 98.6 |

Some sources of false alarms (as seen in Figure 4) are the water from the ceiling alarm (detected as smoke), or very bright light sources (detected as a fire). This issue should be considered further for the system to not be misled by water/mist flows or spotlights. Currently, our method carries out detection on a frame-by-frame basis. A solution to increase the overall accuracy would be to average the output across several frames. This would remove random anomalies (false positives) that may appear at isolated frames while accumulating confidence over time of detections (true positives), even if at some individual frames, their confidence is not so high or are missed (false negatives). To help in this mission, we could also incorporate tracking methods to analyze image regions where previously there was a detection with high accumulated confidence. It should be considered that such solutions would come at the cost of increasing the detection time, although it could be mitigated if the sensor and the system can cope with a sufficient frame rate.

The different performance of fire and smoke detection is also worth studying further by examining the composition of the datasets and maybe training a separate method for each. Since smoke may expand faster than flames, it can provide an earlier clue for the detection task, so improving its performance is highly relevant. Also, newer real-time detectors like YOLOR (Wang et al., 2021) are showing higher speed than YOLOv4 while keeping accuracy, which could favor its deployment to embedded devices in industrial settings. This system was released after the course of this investigation, so it is saved for future work.

When deploying a fire detection system like the one in this paper, one must consider the various ethical questions related to camera-based detection, due to humans potentially appearing in the footage. Whenever a camera is capturing or streaming such type of data to a remote location, privacy, security, and GDPR concerns emerge. These concerns would be significantly counteracted via edge computing, with data processed as close as possible to where it is being captured, diminishing transmission of sensitive data to a different location through data networks. In this regard, edge devices usually have fewer computing capabilities, which is the reason why we are aiming at deploying our system to suitable hardware, such as NVIDIA Jetson nano. This also connects with using detectors with low latency, such as YOLOR, as mentioned above. Also, necessary frames must be deleted as soon as computations are done. The present system only uses one frame, but even combining several frames with sufficient frame rate would mean that the necessary data to be processed only affects a few milliseconds of footage. Handling the data in this way means that no sensitive data would ever be stored, or transmitted elsewhere.

# ACKNOWLEDGEMENTS

This work has been carried out by Otto Zell and Joel Pålsson in the context of their Bachelor Thesis at Halmstad University (Computer Science and Engineering), with the support of HMS Networks AB in Halmstad. Authors Hernandez-Diaz and Alonso-Fernandez thank the Swedish Research Council (VR) and the Swedish Innovation Agency (VINNOVA) for funding their research.

# REFERENCES

Abdusalomov, A., Baratov, N., Kutlimuratov, A., and Whangbo, T. K. (2021). An improvement of the fire detection and classification method using yolov3 for surveillance systems. *Sensors*, 21(19).

Accosta, R. and Martin, D. (2017). Smoke detector spacing for high ceiling spaces. Technical report, National Fire Protection Association, NFPA.

Ahrens, M. (2021). Smoke alarms in us home fires. Technical report, National Fire Protection Association, NFPA.

Avazov, K., Mukhiddinov, M., Makhmudov, F., and Cho, Y. I. (2022). Fire detection method in smart city environments using a deep-learning-based approach. *Electronics*, 11(1).

Bochkovskiy, A., Wang, C.-Y., and Liao, H.-Y. M. (2020). Yolov4: Optimal speed and accuracy of object detection.

Bu, F. and Gharajeh, M. S. (2019). Intelligent and vision-based fire detection systems: A survey. *Image and Vision Computing*, 91:103803.

Fonollosa, J., Solórzano, A., and Marco, S. (2018). Chemical sensor systems and associated algorithms for fire detection: A review. *Sensors*, 18(2).

Gonzalez, A., Zuniga, M. D., Nikulin, C., Carvajal, G., Cardenas, D. G., Pedraza, M. A., Fernandez, C. A., Munoz, R. I., Castro, N. A., Rosales, B. F., Quinteros, J. M., Rauh, F. J., and Akhloufi, M. A. (2017). Accurate fire detection through fully convolutional network. In *7th Latin American Conference on Networked and Electronic Media (LACNEM 2017)*, pages 1–6.

HMS (2022). *https://www.hms-networks.com*.

Li, Z., Peng, C., Yu, G., Zhang, X., Deng, Y., and Sun, J. (2017). Light-head r-cnn: In defense of two-stage object detector.

Muhammad, K., Ahmad, J., Lv, Z., Bellavista, P., Yang, P., and Baik, S. W. (2019). Efficient deep cnn-based fire detection and localization in video surveillance applications. *IEEE Transactions on Systems, Man, and Cybernetics: Systems*, 49(7):1419–1434.

Nilsson, F., Jakobsen, J., and Alonso-Fernandez, F. (2020). Detection and classification of industrial signal lights for factory floors. In *2020 International Conference on Intelligent Systems and Computer Vision (ISCV)*, pages 1–6.

Tan, M., Pang, R., and Le, Q. V. (2020). Efficientdet: Scalable and efficient object detection. In *2020 IEEE/CVF Conference on Computer Vision and Pattern Recognition (CVPR)*, pages 10778–10787, Los Alamitos, CA, USA. IEEE Computer Society.

Wang, C.-Y., Yeh, I.-H., and Liao, H.-Y. M. (2021). You only learn one representation: Unified network for multiple tasks. *arXiv preprint arXiv:2105.04206*.

Wu, D., Lv, S., Jiang, M., and Song, H. (2020). Using channel pruning-based yolo v4 deep learning algorithm for the real-time and accurate detection of apple flowers in natural environments. *Computers and Electronics in Agriculture*, 178:105742.

Zhang, J., Zhu, H., Wang, P., and Ling, X. (2021). Att squeeze u-net: A lightweight network for forest fire detection and recognition. *IEEE Access*, 9:10858–10870.